\documentclass[letterpaper, 10 pt, conference]{ieeeconf}
\usepackage{amsmath,amsfonts}
\usepackage{algorithmic}
\usepackage{algorithm}
\usepackage{array}
\usepackage[caption=false,font=normalsize,labelfont=rm,textfont=rm]{subfig}
\usepackage{textcomp}
\usepackage{stfloats}
\usepackage{url}
\usepackage{verbatim}
\usepackage{graphicx}
\IEEEoverridecommandlockouts
\usepackage{epsfig} 
\usepackage{epstopdf}
\usepackage{courier}

\usepackage{color}

\usepackage{amssymb}
\usepackage{float}
\usepackage[backend=bibtex,bibstyle=ieee,citestyle=numeric-comp]{biblatex}
\DeclareMathOperator*{\argmin}{arg\,min}

\usepackage{siunitx}

\usepackage{balance}
\title{\LARGE \bf
Magnetic Navigation using Attitude-Invariant Magnetic Field Information for Loop Closure Detection
}
        
\author{Natalia Pavlasek$^1$, Charles Champagne Cossette$^1$, David Roy-Guay$^2$, and James Richard Forbes$^1$
\thanks{*This work was supported the Mitacs Accelerate program, IVADO, SBQuantum, the Canadian Foundation of Innovation, NSERC Discovery Grant, and the McGill University MEUSMA program.
}
\thanks{$^{1}$Department of Mechanical Engineering, McGill University, 817 Sherbrooke St. W., Montreal, QC,
Canada, H3A 0C3. (e-mail: {\small natalia.pavlasek@mail.mcgill.ca, charles.cossette@mail.mcgill.ca, james.richard.forbes@mcgill.ca.})}
\thanks{$^{2}$SBQuantum, 805 Galt St. W., Sherbrooke, QC, Canada, J1H 1Z1. (e-mail: {\small david@sbquantum.com.})}
\vspace{-1cm}}

\newcommand{\norm}[1]{\left\Vert#1\right\Vert} 
\newcommand{\mc}[1]{\mathcal{#1}}

\newcommand{\bma}[1]{\left[\begin{array}{ #1}}
\newcommand{\ema}{\end{array}\right]}

\DeclareMathAlphabet{\mbf}{OT1}{ptm}{b}{n}
\newcommand{\mbs}[1]{{\boldsymbol{#1}}}

\newcommand{\mbfbar}[1]{{\bar{\mbf{#1}}}}
\newcommand{\mbfhat}[1]{{\hat{\mbf{#1}}}}

\def\fdotb{{\raisebox{-0.6ex}{ \kern0.2ex\raisebox{0.8ex}{\tiny $\hspace*{-1ex}\circ$}}}}
\def\fddotb{{\raisebox{-0.6ex}{ \kern0.2ex\raisebox{0.8ex}{\tiny $\hspace*{-1ex}\circ\circ$}}}}

\newcommand{\rnums}{\mathbb{R}}

\newcommand{\p}{\partial}
\newcommand{\f}{\frac}

\newcommand{\ura}[1]{{\underrightarrow{{#1}}}}

\newcommand{\trans}{{\ensuremath{\mathsf{T}}}} 
\newcommand{\utimes}{ {\raisebox{-0.6ex}{ \kern-1.0ex\raisebox{0.6ex}{ \small $\mathsf{v}$}}} } %
\newcommand{\beq}{\begin{equation}}
\newcommand{\eeq}{\end{equation}}
\newcommand{\bdis}{\begin{displaymath}}
\newcommand{\edis}{\end{displaymath}}
\newcommand{\beqarray}{\begin{eqnarray}}
\newcommand{\eeqarray}{\end{eqnarray}}
\newcommand{\beqarraynn}{\begin{eqnarray*}}
\newcommand{\eeqarraynn}{\end{eqnarray*}}
\newcommand{\balign}{\begin{align}}
\newcommand{\ealign}{\end{align}}
\newcommand{\balignnn}{\begin{align*}}
\newcommand{\ealignnn}{\end{align}}

\renewcommand{\theenumii}{\arabic{enumii}}
\makeatletter
\renewcommand{\p@enumii}{\theenumi.}
\makeatother

\begin{document}

%
%
%
%
%
%
%
\def \myJournal {IEEE/RSJ International Conference on Intelligent Robots and Systems}
\def \myDoi {}
\def \myPaperSiteName {IEEE Xplore}
\def \myPaperSiteLink {}
\def \myYear {2023}
\def \myPaperCitation{N. Pavlasek, C. Champagne Cossette, D. Roy-Guay, J. Richard Forbes, ``Magnetic Navigation using Attitude-Invariant Magnetic Field Information for Loop Closure Detection,'' \url{http://arxiv.org/abs/2307.XXXXX}, 2023.}


\begin{figure*}[t]

\thispagestyle{empty}
\begin{center}
\begin{minipage}{6in}
\centering
This paper has been accepted for presentation at the \emph{\myJournal}. 
\vspace{1em}

This is the author's version of an article that has, or will be, published in this journal or conference. Changes were, or will be, made to this version by the publisher prior to publication.
\vspace{2em}


Please cite this paper as:

\myPaperCitation

\vspace{15cm}
\copyright \myYear \hspace{4pt}IEEE. Personal use of this material is permitted. Permission from IEEE must be obtained for all other uses, in any current or future media, including reprinting/republishing this material for advertising or promotional purposes, creating new collective works, for resale or redistribution to servers or lists, or reuse of any copyrighted component of this work in other works.

\end{minipage}
\end{center}
\end{figure*}
\newpage
\clearpage
\pagenumbering{arabic}

\maketitle
\thispagestyle{empty}
\pagestyle{empty}

\begin{abstract}

Indoor magnetic fields are a combination of Earth's magnetic field and disruptions induced by ferromagnetic objects, such as steel structural components in buildings. As a result of these disruptions, pervasive in indoor spaces, magnetic field data is often omitted from navigation algorithms in indoor environments. This paper leverages the spatially-varying disruptions to Earth's magnetic field to extract positional information for use in indoor navigation algorithms. The algorithm uses a rate gyro and an array of four magnetometers to estimate the robot's pose. Additionally, the magnetometer array is used to compute attitude-invariant measurements associated with the magnetic field and its gradient. These measurements are used to detect loop closure points. {Experimental results indicate that the proposed approach can estimate the pose of a ground robot in an indoor environment within meter accuracy.}

\end{abstract}

\section{Introduction}

Robots working indoors, such as material handling and cleaning robots, rely on robust indoor navigation solutions. Robots are often equipped with a camera or LiDAR that are used as part of a navigation algorithm. However, a camera can be unreliable when lighting conditions are poor, and LiDAR can be costly and resource-heavy. In recent years, the use of magnetic field measurements for indoor navigation has been studied \cite{Dorveaux2011, Zmitri2019, Vissiere2007, Kok2018, Vallivaara2018, Jung2019, Robertson2013, Frassl2013}. These measurements are attractive from a computational standpoint since the magnetic field is a three dimensional vector field, meaning that the number of measurements to process is small.

In indoor environments, the magnetic field is an amalgamation of the Earth's magnetic field and strong contributions from ferromagnetic and paramagnetic objects. Indoors, environmental features such as structural beams induce a spatially-varying magnetic field, which can be used for localization. Within the field of geophysics, magnetic field gradients have been used in place of magnetic fields to characterize an environment \cite{Heath2003}. The magnetic field gradient can be measured using an array of magnetometers.

\begin{figure}
    \centering
    \includegraphics[width=0.85\columnwidth, trim={6cm 0 6cm 0}, clip]{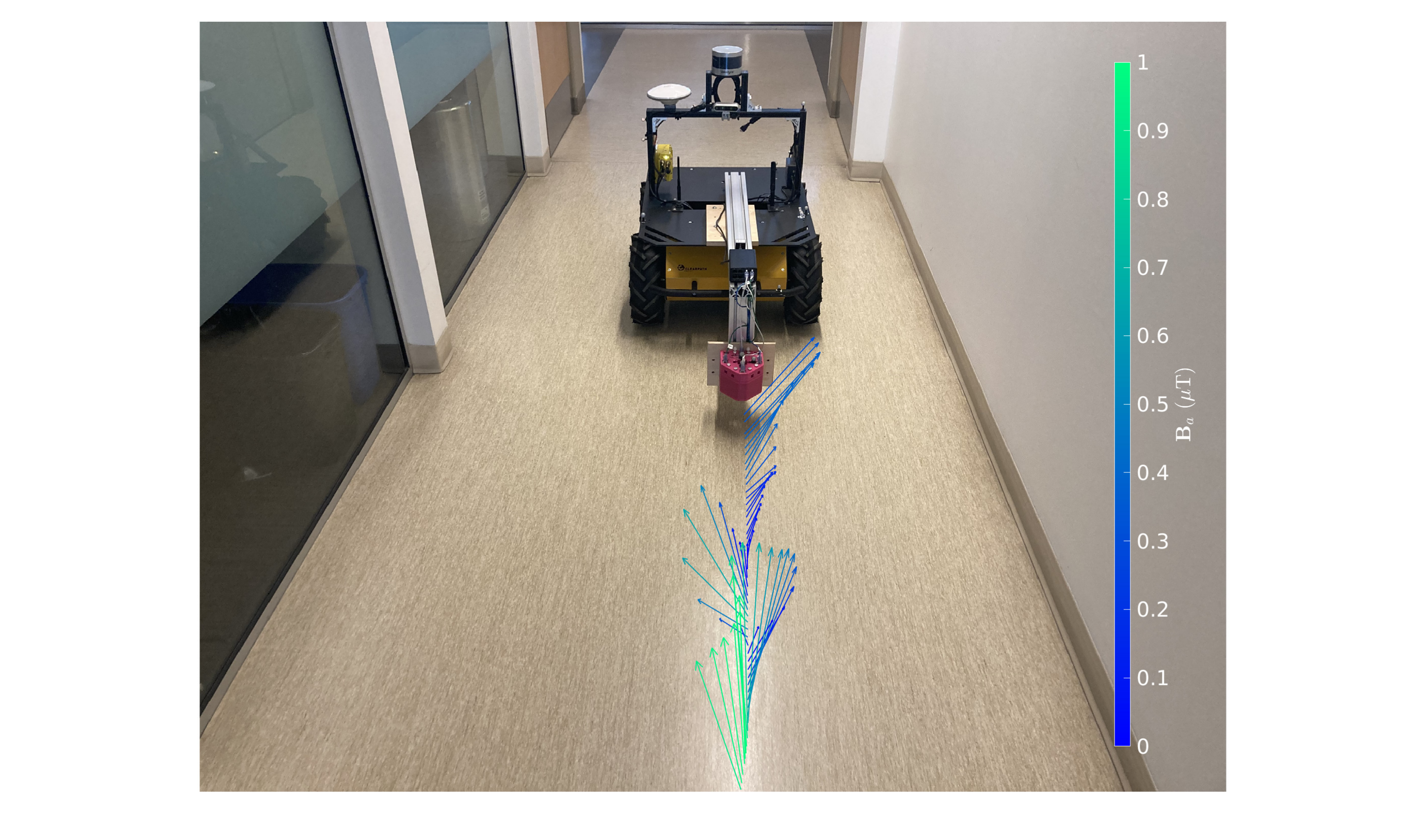}\vspace{-0.5cm}
    \caption{Experiment setup with illustration of magnetic field. The proposed algorithm uses local variations to Earth's magnetic field, induced by ferromagnetic materials in the structure of the building, to navigate.}
    \label{fig:experiment_setup1} \vspace{-0.5cm}
\end{figure}

Loop closures occur when a robot visits a previously visited location and are an important source of information to navigation algorithms. Detecting and enforcing loop closures is important in preventing pose estimates from drifting. In the context of navigation using magnetic fields, detecting and enforcing loop closures has been shown to be critical \cite{Jung2019, Vallivaara2018}.

\subsection{Related Work}

Much of the previous work on magnetic-field-based navigation has used the simultaneous localization and mapping (SLAM) formulation \cite{Kok2018, Vallivaara2018, Jung2019, Robertson2013, Frassl2013, Viset2022, Lee2020}. Though existing techniques have been shown to be effective under certain conditions, they suffer from a lack of robustness and are computationally heavy. The solutions presented in \cite{Viset2022} and \cite{Jung2019} restrict the robot trajectories to those with large overlapping segments. The trajectories are not explicitly restricted in \cite{Kok2018, Vallivaara2018, Frassl2013, Robertson2013}, which use particle-filter-based SLAM. However, the particle-filter-based SLAM approach results in a more computationally demanding solution. In all cases, a surface is fit to the magnetic field measurements. The number of degrees of freedom used to represent the magnetic field has a direct impact on the accuracy of the map. Should the number of degrees of freedom of the map be poorly selected, the magnetic field data can be overfit or magnetic features can be left out of the map.

Magnetic field gradients are used in conjunction with magnetic field measurements in the context of localization within an existing map in \cite{Canciani2016, Mount2018}, and in the context of navigation in \cite{Dorveaux2011, Zmitri2019, Vissiere2007}. It is found in \cite{Mount2018} that localization results experience significant improvement when magnetic field gradient measurements are included in addition to magnetic field measurements.

Loop closure detection and enforcement has been shown to be critical in previous magnetic-field-based navigation solutions. The difference between the norms of any two magnetic field vector measurements and the distance between the two estimated positions at which the measurements were taken are computed in \cite{Vallivaara2010}. These two quantities are weighted in order to detect loop closure points. Loop closures are detected if the norm of the magnetic field vector is the same for multiple subsequent poses in \cite{Jung2019}. Loop closures are therefore detected only if a segment of the trajectory overlaps. Robust loop closure techniques are used to reject possible false loop closure detections.

\subsection{Contributions}

The novel contribution of this work is two-fold. First, a navigation strategy that uses only a rate gyro and an array of four magnetometers is developed. The algorithm uses magnetic field and magnetic field gradient measurements to estimate relative positions between subsequent robot poses. Unlike \cite{Kok2018, Vallivaara2018, Jung2019, Robertson2013, Frassl2013, Viset2022}, map states are not simultaneously estimated with the robot poses, leading to an analytically and numerically simpler navigation solution. {For example, in addition to robot pose states, which are elements of $\rnums^6$, 1536 map states were estimated in \cite{Kok2018}}. Nevertheless, the proposed approach can facilitate extraction of a map. For example, using a surface fitting algorithm, a surface can be fit to the magnetic field measurements at the estimated robot poses.  Avoiding the explicit estimation of maps states is made possible by the availability of magnetic field gradient measurements. The second contribution is the development of a simple, map-free, loop closure detection algorithm. The loop closure detection only requires storing a history of three attitude-invariant scalar quantities that are calculated directly from the magnetic field and magnetic field gradient measurements. The attitude-invariance of these measurements make them well-suited for loop closure detection since potentially large attitude-estimation error does not affect the detection of loop closures. The proposed magnetic-based navigation method is validated in two experiments involving a ground robot endowed with a rate gyro and a four-magnetometer array. A representation of the problem is shown in Figure~\ref{fig:experiment_setup1} where the Clearpath Husky robot is shown in the test environment for one of the experiments with an illustration of the magnetic field.

The proposed approach holds the greatest similarity to the approaches in \cite{Dorveaux2011, Zmitri2019, Vissiere2007}. Each of \cite{Dorveaux2011, Zmitri2019, Vissiere2007} use an accelerometer and a rate gyro in conjunction with four magnetometers. Moreover, in \cite{Dorveaux2011, Zmitri2019, Vissiere2007} the magnetic field measurements are estimated as part of the state. In contrast, the approach proposed herein only relies on a rate gyro and four magnetometers, and only the robot pose is part of the state being estimated. As a result, the proposed approach relies on one fewer sensors, and fewer states are estimated, relative to  \cite{Dorveaux2011, Zmitri2019, Vissiere2007}. In addition, loop closures are not detected in \cite{Dorveaux2011, Zmitri2019, Vissiere2007}.

Note that the use of other sensors is deliberately omitted to assess the effectiveness of pure gyro-magnetic navigation. However, additional sensors can be incorporated in a straightforward manner.

The remainder of this work is organized as follows. Section~\ref{sec:prelim} offers preliminary information on magnetostatics and attitude-invariant measurements. The proposed magnetic navigation algorithm is outlined in Section~\ref{sec:MO}. The loop closure detection algorithm is presented in Section~\ref{sec:loop_closing}. Section~\ref{sec:results} presents the results of the experiments. Finally, concluding remarks and directions for future work are given in Section~\ref{sec:conclusion}.

\section{Preliminaries} \label{sec:prelim}

\subsection{Notation}

The following notation is used in this paper. A three-dimensional physical vector quantity $\ura{p}$ resolved in reference frame $\mc{F}_a$ is denoted $\mbf{p}_a$. The same quantity resolved in $\mc{F}_b$ is denoted $\mbf{p}_b$. A direction cosine matrix (DCM), often referred to as a rotation matrix, is denoted $\mbf{C}_{ab} \in SO(3)$, where the relation between $\mbf{p}_a$ and $\mbf{p}_b$ is ${\mbf{p}_a = \mbf{C}_{ab} \mbf{p}_b}$. Moreover, a generic DCM can be written as ${\mbf{C} = \exp(\mbs{\phi}^\times)}$ where ${\left( \cdot \right)^\times : \mathbb{R}^3} \to \mathfrak{so}(3)$ is the skew-symmetric cross product operator \cite{Barfoot2017,paper_Sola_eta_al_micro}.

\subsection{Magnetostatics}
\label{sec:mag_fields}

Let $\mbf{B}_b(\mbf{r}_b) \in \mathbb{R}^3$ denote the spatially-varying magnetic field vector, where $\mbf{r}_b = \left[ x_b \; y_b \; z_b \right]^\trans$ is the position of the magnetometer array, resolved in the body frame $\mc{F}_b$. For the purpose of this paper, the problem of interest is that of magnetostatics, in which currents and electric fields are ignored. In this case, Maxwell's equations state that the magnetic field is a curl- and divergence-free vector field, meaning that \cite{Purcell2012}
\begin{align*}
    \mbs{\nabla}_b^\trans \mbf{B}_b(\mbf{r}_b) = 0, \quad \quad \mbs{\nabla}_{b}^\times \mbf{B}_b(\mbf{r}_b) = \mbf{0} , \label{eq:maxwell}
\end{align*}
where $\mbs{\nabla}_b = \left[\f{\p}{\p x_b} \; \f{\p}{\p y_b} \; \f{\p}{\p z_b} \right]^\trans$. The magnetic field vector is also resolved in the body frame, since a magnetometer measures $\mbf{B}_b(\mbf{r}_b)$ directly. The curl- and divergence-free properties of the magnetic field result in the gradient of the magnetic field being symmetric and traceless. The magnetic field gradient $\mbf{G}_b(\mbf{r}_b)$ can therefore be represented by five unique elements as
\bdis 
\mbf{G}_b = \frac{\p \mbf{B}_b (\mbf{r}_b)} {\p \mbf{r}_b} = \bma{ccc}
        B_{x_b x_b} & B_{x_b y_b} & B_{x_b z_b}\\
        B_{x_b y_b} & B_{y_b y_b} & B_{y_b z_b}\\
        B_{x_b z_b} & B_{y_b z_b} & -(B_{x_b x_b} + B_{y_b y_b})
    \ema,
\edis
where the argument of $\mbf{G}_b(\mbf{r}_b)$ is omitted for brevity.

The mapping from the magnetic field gradient to a column vector containing the five unique elements of the gradient is defined as $(\cdot)^\vdash : \mathbb{R}^{3\times 3} \rightarrow \mathbb{R}^{5}$, such that
\begin{align*}
    \mbf{G}_b^\vdash = \left[\;
        B_{x_bx_b} \;
        B_{x_by_b} \;
        B_{x_bz_b}\;
        B_{y_by_b}\;
        B_{y_bz_b}\;
    \right]^\trans.
\end{align*}

In practice, the unique components of the gradient can be computed using an array of three magnetometers, and applying a forward finite-difference approximation to compute the derivatives. Using an array of four magnetometers, the derivatives in the two planar directions are centered and collocated. A detailed discussion of magnetometer calibration is outside of the scope of this paper but is outlined in \cite{Dorveaux2011}.

\subsection{Attitude-invariant Measurements}\label{sec:invariants}

The magnetic field measurement, $\mbf{B}_b$, and the magnetic field gradient measurement, $\mbf{G}_b$, can be used to define three scalar values, referred to as invariants owing to their invariance to changes in attitude. The three invariants are
\begin{align}
    I_1 = \norm{\mbf{B}_b}_2 , \quad \quad
    I_2 = \norm{\mbf{G}_b}_\mathsf{F} , \quad \quad
    I_3 = \det (\mbf{G}_b) \label{eq:I1I2I3} .
\end{align}
The first invariant, $I_1$, is the Euclidean norm of $\mbf{B}_b$. Notice ${I_1 = \norm{\mbf{B}_b}_2 = \norm{\mbf{C}_{bc} \mbf{B}_c}_2 = \norm{\mbf{B}_c}_2}$, where $\mbf{B}_c$ is the magnetic field measurement resolved in another arbitrary body frame $\mc{F}_c$. Thus, $I_1$ is invariant to changes in robot attitude. The second invariant is the Frobenius norm of $\mbf{G}_b$. Notice that ${I_2 = \norm{\mbf{G}_b}_\mathsf{F} = \norm{\mbf{C}_{bc} \mbf{G}_c \mbf{C}_{bc}^\trans}_\mathsf{F} = \norm{\mbf{G}_c}_\mathsf{F}}$, where $\mbf{G}_c$ is the magnetic field gradient measurement resolved in $\mc{F}_c$. Like $I_1$, $I_2$ is invariant to attitude changes. The third invariant is the determinant of $\mbf{G}_b$. Notice that ${I_3 = \det (\mbf{G}_b) = \det ( \mbf{C}_{bc} \mbf{G}_c \mbf{C}_{bc}^\trans ) = \det ( \mbf{C}_{bc} ) \det ( \mbf{G}_c ) \det ( \mbf{C}_{bc}^\trans )}$ ${ = \det ( \mbf{G}_c )}$, where $\det ( \mbf{C}_{bc} ) = \det ( \mbf{C}_{bc}^\trans ) = +1$. As with $I_1$ and $I_2$, $I_3$ is invariant to changes in attitude.

The attitude invariant nature of $I_1$, $I_2$, and $I_3$ is attractive from a navigation standpoint because magnetic field information can be measured and mapped as a function of the position only. In this paper, these three invariants will be used to detect loop closures, as discussed in Section~\ref{sec:loop_closing}.

\section{Magnetic Navigation} \label{sec:MO}
This section describes the problem of estimating robot poses using rate-gyro, magnetic field, and magnetic field gradient measurements. The proposed approach relates magnetic field and magnetic field gradient measurements to the robot pose. As such, the poses can be estimated using only rate-gyro, magnetic field, and magnetic field gradient measurements. Unlike \cite{Kok2018, Vallivaara2018, Jung2019, Robertson2013, Frassl2013, Viset2022}, where the magnetic field map is parameterized and map states are explicitly estimated, the proposed approach eliminates the need to estimate map states. This significantly reduces the size of the problem compared to approaches such as \cite{Kok2018}, which estimates 1536 map states in addition to robot states. Additionally, while some of the existing approaches impose restrictions on the trajectory \cite{Viset2022, Jung2019}, the proposed approach does not require these restrictions.

\subsection{Rate-gyro Model} \label{sec:problem}
The rate-gyro measurement at timestep $k$ is represented as
\begin{align*}
	u_{b_k} &= \omega_{b_k}^{ba} - w_{b_k},
\end{align*}
where $u_{b_k}$ is the rate-gyro measurement, $\omega_{b_k}^{ba}$ is the angular velocity of $\mc{F}_b$ relative to an inertial frame $\mc{F}_a$ resolved in $\mc{F}_b$, and ${w_{b_k} \sim \mathcal{N}(0,\mathrm{Q}_k)}$ is zero-mean Gaussian white measurement noise. This measurement is a scalar for ground robots. The rate gyro contributes attitude information through the model
\begin{align}
    \mbf{C}_{ab_{k+1}} = \mbf{C}_{ab_k} \exp\left( \Delta t \left( u_{b_k} + w_{b_k} \right)^\times \right), \label{eq:rate_gyro}
\end{align}
where $\Delta t$ is the time between two rate-gyro measurements, and for 2D DCMs $(\cdot)^\times: \mathbb{R} \to \mathfrak{so}(2)$ \cite{Barfoot2017}.

\subsection{Magnetic-field Pseudomeasurement}
An array of four magnetometers is used to measure the magnetic field and its gradient. The resulting measurements at timestep $k$ are
\begin{align*}
	\mbf{y}_k^\mbf{B} &= \mbf{B}_{b_k} + \mbf{v}^\mbf{B}_k \in \mathbb{R}^3, \quad \quad \mbf{v}^\mbf{B}_k \sim \mathcal{N}(\mbf{0}, \mbf{R}^\mbf{B}_k),\\
	\mbf{y}_k^\mbf{G} &= \mbf{G}^\vdash_{b_k} + \mbf{v}^\mbf{G}_k \in \mathbb{R}^5,  \quad \quad \mbf{v}^\mbf{G}_k \sim \mathcal{N}(\mbf{0}, \mbf{R}^\mbf{G}_k).
\end{align*}
Positional information is extracted from the magnetic field and magnetic field gradient measurements through the introduction of a pseudomeasurement. A pseudomeasurement is a purely-synthetic measurement whose value is exactly zero.

\begin{figure}
    \begin{center}$
        \begin{array}{cccc}
            \includegraphics[width=30mm]{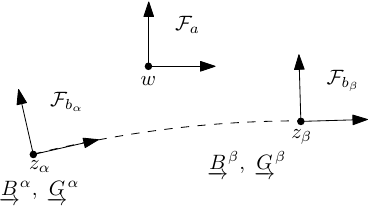}&&&
            \includegraphics[width=17mm]{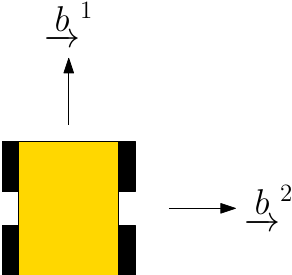}
        \end{array}$
    \end{center}
    \caption{\emph{Left}: Illustration of three subsequent robot body frames. Quantities $\ura{B}^k$ and $\ura{G}^k$, where $k \in \{\alpha, \beta, \gamma\}$, are the magnetic field and magnetic field gradient, respectively, measured at each point. \emph{Right}: Illustration of top view of Husky robot with body frame basis vectors.}
    \label{fig:frames}\vspace{-0.5cm}
\end{figure}

Consider a first-order Taylor series expansion of $\mbf{B}_b(\mbf{r}_b)$ about an arbitrary nominal position $\mbfbar{r}_b$
\begin{align}
	\mbf{B}_b(\mbf{r}_b) \approx \mbf{B}_b(\mbfbar{r}_b) + \mbf{G}_b(\mbfbar{r}_b) (\mbf{r}_b -  \mbfbar{r}_b). \label{eq:TSE}
\end{align}
Consider subsequent timesteps, labelled $\alpha,\; \beta$. Let $\mbf{B}_{b_i}^j$ be the magnetic field at timestep $j$, resolved in the body frame at timestep $i$, $\mathcal{F}_{b_i}$, where $i, j \in \left\{ \alpha, \beta \right\}$. Similarly, $\mbf{G}_{b_i}^j$ is the magnetic field gradient at timestep $j$ resolved in $\mathcal{F}_{b_i}$. This problem is illustrated in Figure~\ref{fig:frames}. Equation \eqref{eq:TSE} can be applied, with $\mbf{r}_b = \mbf{r}_{b_\beta}^{z_\beta w}$ and $\mbfbar{r}_b = \mbf{r}_{b_\beta}^{z_\alpha w}$ being the position from stationary point $w$ to points $z_\beta$ and $z_\alpha$, respectively, resolved in $\mathcal{F}_{b_\beta}$. The result is
\begin{align}
    \mbf{B}_{b_\beta}^\beta - \mbf{B}_{b_\beta}^\alpha \approx \mbf{G}_{b_\beta}^\beta (\mbf{r}_{b_\beta}^{z_\beta w} - \mbf{r}_{b_\beta}^{z_\alpha w}), \label{eq:gradient_dt}
\end{align}
where all quantities are resolved in $\mathcal{F}_{b_\beta}$. Equation \eqref{eq:gradient_dt} is a forward-difference-like approximation at timestep $\beta$. Resolving the measured quantities in \eqref{eq:gradient_dt} in the frames in which they are measured gives
\begin{align*}
    \mbf{B}_{b_\beta}^\beta - \mbf{C}_{ab_\beta}^\trans \mbf{C}_{ab_\alpha}\mbf{B}_{b_\alpha}^\alpha \approx \mbf{G}_{b_\beta}^\beta \mbf{C}_{ab_\beta}^\trans( \mbf{r}_{a}^{z_\beta w} - \mbf{r}_{a}^{z_\alpha w}), 
\end{align*}
which finally motivates the following pseudomeasurement
\begin{align}
    \mbf{y}_\beta^{\mathrm{FD}} &= \mbf{G}_{b_\beta}^\beta \mbf{C}_{ab_\beta}^\trans \left( \mbf{r}_a^{z_\beta w} - \mbf{r}_a^{z_\alpha w} \right) \nonumber \\
    &\hspace{8mm}- \left( \mbf{B}_{b_\beta}^\beta - \mbf{C}_{ab_\beta}^\trans \mbf{C}_{ab_\alpha}\mbf{B}_{b_\alpha}^\alpha \right) + \mbf{v}_\beta^{\mathrm{FD}}, \label{eq:pseudo2}
\end{align}
where $\mbf{v}_\beta^{\mathrm{FD}} \sim \mathcal{N}(\mbf{0}, \mbf{R}_\beta^{\mathrm{FD}})$, and $\mbf{R}_\beta^{\mathrm{FD}} \in \mathbb{R}^{3\times 3}$. This pseudomeasurement is always ``measured'' to be exactly zero, yet the approximation in \eqref{eq:gradient_dt} only holds to first order, which is an inaccuracy accounted for by the noise term $\mbf{v}_\beta^{\mathrm{FD}}$.

Similarly, a central-difference-like pseudomeasurement can be created in an identical manner as before using timesteps $\alpha$, $\beta$, and $\gamma$, leading to 
\begin{align}
    \mbf{y}_\beta^{\mathrm{CD}} &= \mbf{G}_{b_\beta}^\beta \mbf{C}_{ab_\beta}^\trans \left( \mbf{r}_a^{z_\gamma w} - \mbf{r}_a^{z_\alpha w} \right) \nonumber \\
    &\hspace{8mm}- \left( \mbf{C}_{ab_\beta}^\trans \mbf{C}_{ab_\gamma} \mbf{B}_{b_\gamma}^\gamma - \mbf{C}_{ab_\beta}^\trans \mbf{C}_{ab_\alpha}\mbf{B}_{b_\alpha}^\alpha \right) + \mbf{v}_\beta^{\mathrm{CD}}, \label{eq:pseudo3}
\end{align}
where $\mbf{v}_\beta^{\mathrm{CD}} \sim \mathcal{N}(\mbf{0}, \mbf{R}_\beta^{\mathrm{CD}})$, and $\mbf{R}_\beta^{\mathrm{CD}} \in \mathbb{R}^{3\times 3}$. One would expect numerical errors resulting from~\eqref{eq:pseudo3} to be smaller than those from~\eqref{eq:pseudo2}, since a central-difference approximation is more accurate than a forward-difference approximation. However, it turns out that using only \eqref{eq:pseudo3} within a navigation pipeline yields poor estimation performance due to the fact that \eqref{eq:pseudo3} relates poses two timesteps apart, between timesteps $\alpha$ and $\gamma$, causing a decoupling between the odd-numbered poses and even-numbered poses. Therefore, the two pseudomeasurements are used in conjunction.

Since the magnetic field and its gradient are measured in three dimensions, the positions and DCMs in \eqref{eq:pseudo3}, \eqref{eq:pseudo2}, must be represented in three dimensions. If only 2D poses are estimated, the position vectors and attitude DCMs should be ``padded'' appropriately to form 3D quantities, exclusively for the evaluation of \eqref{eq:pseudo3}, \eqref{eq:pseudo2}.

\subsection{Wheel-slip Constraint}
The ground vehicle modelled in this paper is assumed to have no lateral velocity. As such, an additional pseudomeasurement can be included to account for the fact that little motion is expected to occur in the $\ura{b}^2$ direction, illustrated on the right side of Figure~\ref{fig:frames}.

The wheel-slip pseudomeasurement is given by \cite{Brossard2020}
\begin{align}
    {y}_\beta^{\mathrm{slip}} &= \bma{cc} 0 & 1 \ema \left( \mbf{C}_{ab_\beta}^\trans \left( \mbf{r}_a^{z_\beta w} - \mbf{r}_a^{z_\alpha w} \right)\right) + v_\beta^{\mathrm{slip}} , \label{eq:slip}
\end{align}
where $v_\beta^{\mathrm{slip}}$ is zero-mean Gaussian white noise with covariance $R_\beta^{\mathrm{slip}}$. The pseudomeasurement ``measures'' zero lateral velocity in the body frame, and includes noise to account for the fact that some slip can occur.

\subsection{Nonlinear Weighted Least Squares}
The navigation problem is solved in a batch framework using nonlinear weighted least squares \cite{Barfoot2017}. Errors are defined for the rate-gyro model in \eqref{eq:rate_gyro}, and the pseudomeasurements in \eqref{eq:pseudo3}-\eqref{eq:slip}. Note that the pseudomeasurement in \eqref{eq:pseudo2} is essential to the proposed algorithm since it relates subsequent poses. Using \eqref{eq:pseudo3} alone increases the error, since each robot position is only related to every second robot position. It was found that using the pseudomeasurement \eqref{eq:pseudo3} in addition to \eqref{eq:pseudo2} generated the best results.

Following \cite{Barfoot2017}, a batch state estimation problem is formed using pose matrices $\mbf{T}_k \in SE(2)$ to represent the state. Details and closed-form expressions for the various operations associated with $SE(2)$ can be found in \cite{Barfoot2017,paper_Sola_eta_al_micro}. The prior, process, and measurement errors are defined as, and distributed according to,
\bdis\arraycolsep=0.8pt\def\arraystretch{2.1}
\begin{array}{rll}
    \mbf{e}_0 &= \log \left(\mbf{T}_0^{-1} \check{\mbf{T}}_0 \right)^\vee &\sim \mc{N}(\mbf{0}, \mbf{P}_0),\\
    {e}_{u_k} &=  \log \left(\mbf{C}_{ab_k}^{-1} \mbf{C}_{ab_{k-1}}, \exp(\Delta t {u}_{b_{k-1}}^\times) \right)^\vee&\sim \mc{N}(0, \mathrm{Q}_k),\\
    \mbf{e}_{y_k} &= \mbf{y}_k = \left[\mbf{y}_k^\mathrm{CD} \;\; \mbf{y}_k^{\mathrm{FD}} \;\; {y}_k^{\mathrm{slip}} \right]^\trans&\sim \mc{N}(\mbf{0}, \mbf{R}_k),
\end{array}
\edis
respectively. The term $\check{\mbf{T}}_0$ is a prior ``mean'' with covariance $\mbf{P}_0$, and ${\mbf{R}_k = \mathrm{diag}( \mbf{R}_k^\mathrm{CD},\; \mbf{R}_k^\mathrm{FD},\; {R}_k^\mathrm{slip} )}$. Closed-form expressions for the $(\cdot)^\vee$ operator associated with $SE(2)$ and $SO(2)$ can be found in \cite{paper_Sola_eta_al_micro}. Denoting ${\mbf{T} = \left(\mbf{T}_0, \mbf{T}_1, \ldots, \mbf{T}_K\right)}$ as the robot pose trajectory, a nonlinear weighted least squares problem can be constructed as 
\beq 
\mbfhat{T} = \argmin_{\mbf{T}} \; \frac{1}{2} \norm{\mbf{e}_0}^2_{\mbf{P}_0}  + \frac{1}{2} \sum_{k = 1}^K \left( \norm{e_{u_k}}^2_{\mathrm{Q}_k} + \norm{\mbf{y}_{y_k}}^2_{\mbf{R}_k}  \right), \label{eq:batch_min2}
\eeq 
where $\norm{\mbf{v}}^2_{\mbf{S}} = \mbf{v}^\trans \mbf{S}^{-1} \mbf{v}$ denotes the squared Mahalanobis distance. The solution to \eqref{eq:batch_min2} is found using the Gauss-Newton method solved using an on-manifold nonlinear weighted least squares solver \cite{Barfoot2017, Ceres}.

\section{Loop Closure} \label{sec:loop_closing}

\begin{figure}
\begin{center}$
\begin{array}{cc}
\subfloat[{\footnotesize $\textit{I}_1$}]{\includegraphics[width=0.45\columnwidth, trim={1cm 0 1.7cm 0.7cm}, clip]{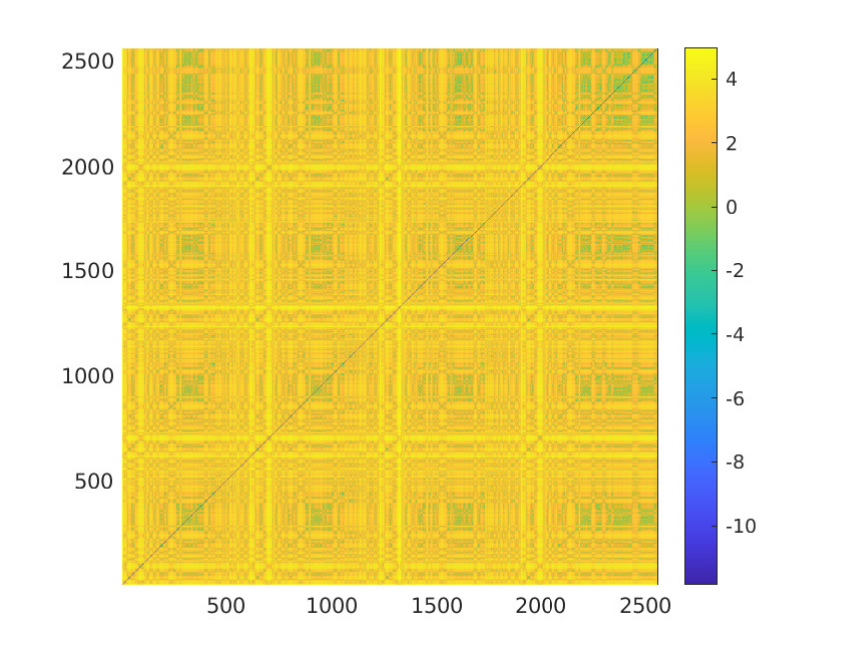}\label{fig:dist_I1}}&
\subfloat[{\footnotesize $\textit{I}_2$}]{\includegraphics[width=0.45\columnwidth, trim={1cm 0 1.7cm 0.7cm}, clip]{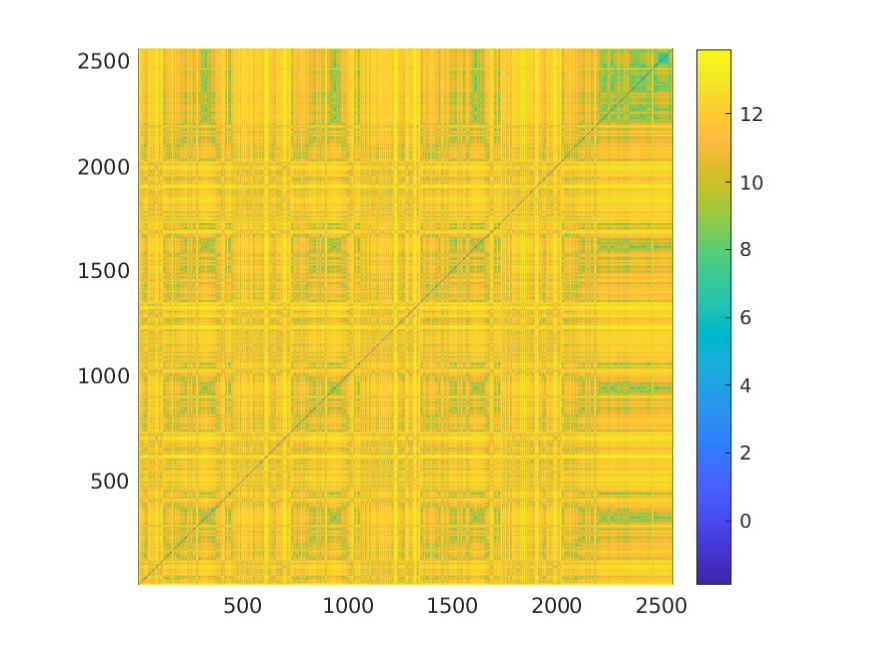}\label{fig:dist_I2}}\\
\subfloat[{\footnotesize $\textit{I}_3$}]{\includegraphics[width=0.45\columnwidth, trim={1cm 0 1.7cm 0.7cm}, clip]{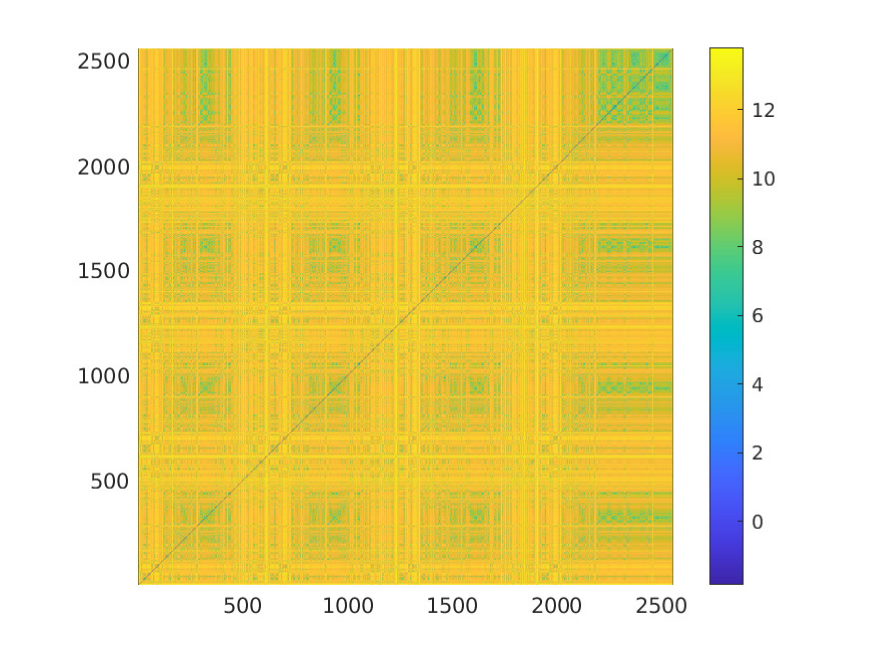}\label{fig:dist_I3}}&
\subfloat[{\footnotesize Sum of normalized invariants}]{\includegraphics[width=0.45\columnwidth, trim={1cm 0 1.7cm 0.7cm}, clip]{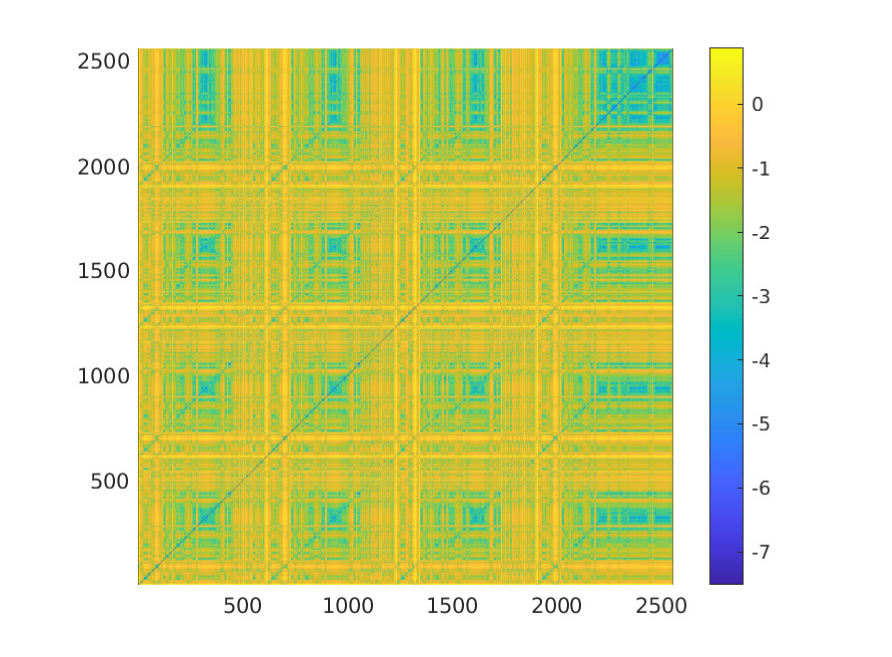}\label{fig:dist_sum}}\\
\end{array}$
\end{center}
\caption{Distance matrices for loop closure detection throughout a trajectory. Axes represent the measurement number and colour represents distance between measurements. The grid-like appearance of these plots stems from the repeating nature of the trajectory of Experiment 1.}
\label{fig:dist} \vspace{-0.5cm}
\end{figure}

The invariants $I_1$, $I_2$, $I_3$ in \eqref{eq:I1I2I3} are used to identify  loop closure candidates. For each measurement, the difference between the measurement and all other measurements is computed, where the distance is $d_{\beta, k \ell} = | I_{\beta,k} - I_{\beta,\ell}|$, $\beta = 1,2,3$, and $I_{\beta,k}$ denotes invariant $\beta$ evaluated at time $k$. This is performed for each of the three invariants, and can be visualized as the matrices shown in Figure~\ref{fig:dist_I1}-\ref{fig:dist_I3}, where the axes represent the measurement number, and the colour indicates the logarithm of the difference between the measured value of the invariants. 

From only one of the invariants, it would be very difficult to detect loop closures. The three distance matrices for the three invariants are normalized by the maximum value of the relevant invariant and summed to give the distance matrix shown in Figure~\ref{fig:dist_sum}. Because the invariants contain sufficiently varied information, it becomes possible to detect the loop closure points as points with small values in the distance matrix.

With the addition of loop closures, the optimization problem in \eqref{eq:batch_min2} gains additional error terms,
\bdis 
\frac{1}{2}\sum_{n = 1}^N \norm{\mbf{e}_{LC_n}}_{\mbs{\Psi}}^2, \qquad \mbf{e}_{LC_n} = \mbf{r}_a^{z_jw} - \mbf{r}_a^{z_iw},
\edis
where $N$ is the number of loop closures, and $i$ and $j$ are the indices of the two poses at which the loop closure occurs. The loop closure error has covariance $\mbs{\Psi}$, which is a fixed value that is initialized using the covariance of the magnetometers and inflated to account for discrepancies caused by the discrete nature of the measurements.

\begin{figure}
    \centering
    \includegraphics[width=0.7\columnwidth]{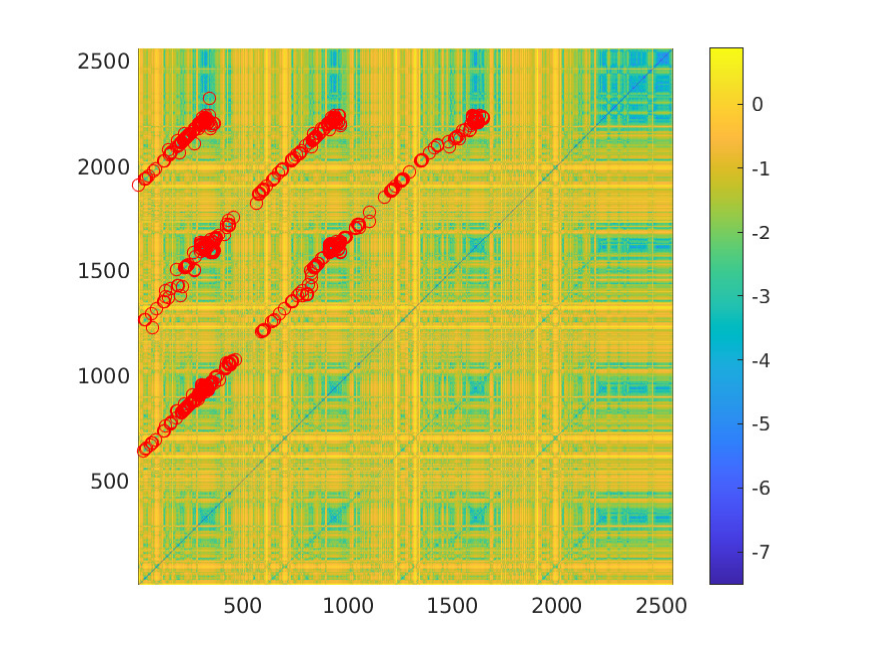}\vspace{-10pt}
    \caption{Distance matrix for loop closure detection with loop closure points detected by the algorithm circled in red. Though the distance matrix is symmetric, loop closure points are only shown on one half of the distance matrix to avoid redundant copies.}
    \label{fig:dist_sum_LC} \vspace{-0.5cm}
\end{figure}

If the magnetic field does not have sufficient variation, or has repeating signals, false loop closure points may be detected. This is apparent in Figure~\ref{fig:dist_sum}, where there appear to be several potential loop closure points. For this reason, the Mahalanobis distance is used to eliminate loop closure points that are infeasible \cite{Neira2001}. A loop closure between states $i$ and $j$ is feasible if
\begin{align*}
    \delta \mbs{\xi}_{ij}^\trans  \mbs{\Sigma}_{ij}^{-1} \delta \mbs{\xi}_{ij} \leq \chi_{d,\alpha}^2, \qquad \delta \mbs{\xi}_{ij} &= \log (\mbf{T}_i^{-1} \mbf{T}_j)^\vee,
\end{align*}
where $\delta \mbs{\xi}_{ij}$ is the ``distance'' between states $i$ and $j$, $\mbs{\Sigma}_{ij}$ is the relative covariance between the two states, computed according to \cite{Mangelson2020}, and $\chi_{d, \alpha}^2$ is the critical Chi-square value for $d$ degrees of freedom and significance level $\alpha$. A significance level of 0.05 is selected. The loop closure points detected using the method detailed herein are circled in red in Figure~\ref{fig:dist_sum_LC}.

\section{Experimental Results} \label{sec:results}
\begin{figure}
    \centering
    \includegraphics[width=0.6\columnwidth]{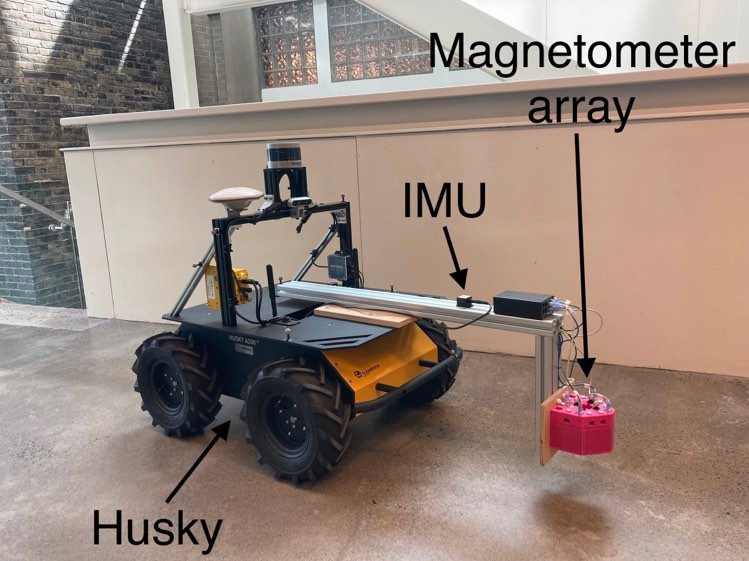}
    \caption{Experiment setup. An array of four magnetometers is mounted to the Clearpath Husky using an aluminum boom.}
    \label{fig:experiment_setup} \vspace{-0.5cm}
\end{figure}

The proposed magnetic  navigation algorithm is tested in two experiments performed on a Clearpath Husky ground robot. The Husky is equipped with an IMU used to collect rate-gyro data. An array of four Twinleaf vector magnetometers \cite{Twinleaf} is used to compute the magnetic field and its gradients. The test setup is shown in Figure~\ref{fig:experiment_setup}.

Data is collected in a hallway and in a laboratory space, both at McGill University. The magnetometer array is placed on a boom to reduce the contribution of the magnetic signature of the Husky to the measured magnetic field. The Husky is also used to collect wheel odometry, which is exclusively used to compute a trajectory estimate to which the proposed algorithm is compared. During the experiments, the velocity of the Husky varies between 0.5 and $1.0 \, \frac{ \SI{}{\meter}}{ \SI{}{\second}}$.

\subsection{Experiment 1}

The first experiment is performed in a hallway, in a space measuring approximately $16\times 27 \, \SI{}{\meter}^2$. The covariance of the measurements is found through Monte Carlo sampling. The covariance of the pseudomeasurement in \eqref{eq:pseudo3} is ${\mbf{R}_k^{\mathrm{CD}} = \left( 5 \times 10^{-1} \, \SI{}{\micro\tesla} \right)^2 \times \mbf{1}}$. The pseudomeasurement in \eqref{eq:pseudo2} has covariance $\mbf{R}_k^{\mathrm{FD}} = \left( 5 \times 10^{0} \, \SI{}{\micro\tesla}\right)^2 \times \mbf{1}$, and the pseudomeasurement in \eqref{eq:slip} has covariance $\mbf{R}_k^{\mathrm{slip}} = \left( 1\times 10^{-4} \, \SI{}{\meter} \right)^2 \times \mbf{1}$. Note that the lower the standard deviation, the greater the effect on correcting the state estimate. Physically, this small standard deviation is equivalent to stating that the ground robot has almost zero side-slip velocity. The loop closure covariance is $\mbs{\Psi}_n = \left( 3.5\times 10^{0} \, \SI{}{\meter} \right)^2 \times \mbf{1}$. The rate gyro has covariance $\mathrm{Q}_k = \left( 1.3\times 10^{-1} \, \frac{\SI{}{\radian}}{\sqrt{\SI{}{\second}}} \right)^2$. The magnetic field and magnetic field gradient measurements are down-sampled to $5 \, \SI{}{\hertz}$. IMU preintegration is used to assimilate the rate gyro and magnetometer frequencies \cite{preint2017}.

\begin{figure}
    \centering
    \includegraphics[width=0.8\columnwidth]{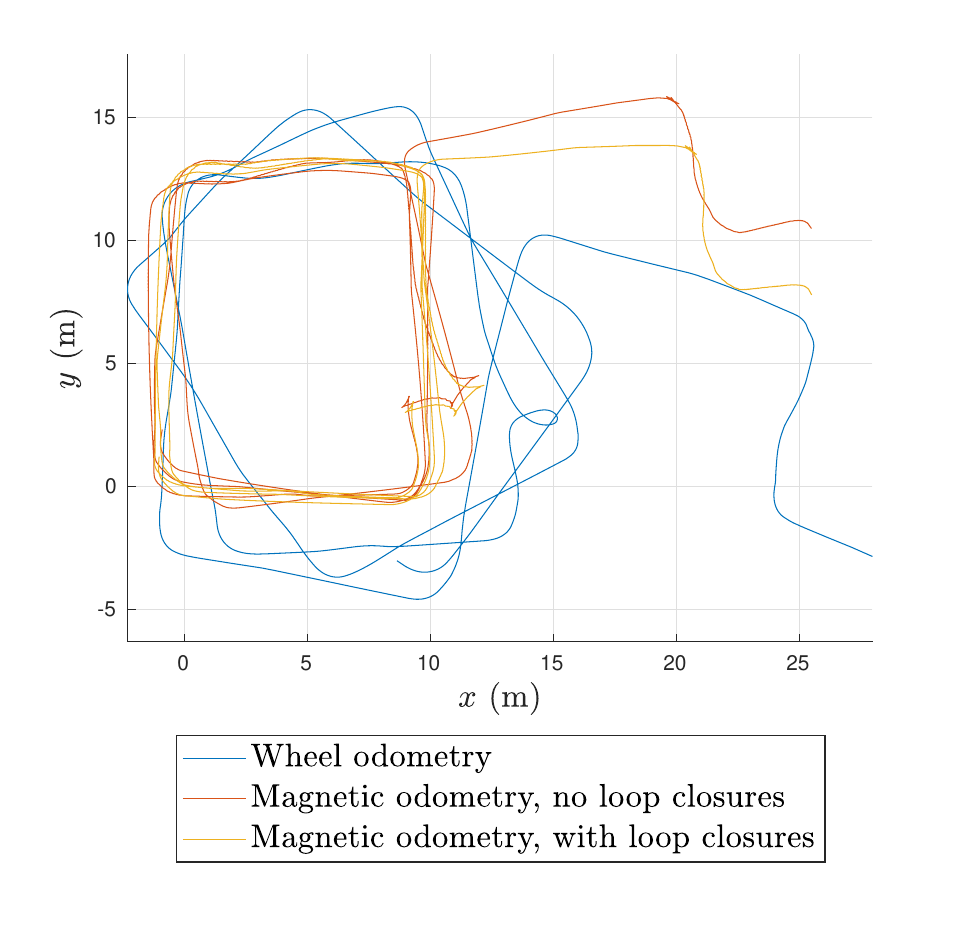}\vspace{-10pt}
    \caption{Top view of trajectories estimated in Experiment 1. The blue trajectory shows the solution estimated using wheel odometry and rate-gyro measurements. The orange trajectory uses the proposed solution without enforcing loop closures. The yellow trajectory is estimated using the proposed solution with loop closures.}
    \label{fig:birds_eye} \vspace{-0.5cm}
\end{figure}

The estimated trajectories are presented in Figure~\ref{fig:birds_eye}. The trajectory represented by the blue line is the solution dead reckoned using wheel odometry and rate-gyro measurements. The orange line shows the trajectory estimated using the proposed solution, without enforcing loop closures. Finally, the yellow line shows the trajectory estimated using the proposed solution with loop closing. The magnetic navigation solutions with and without enforcing loop closures outperform the solution using wheel odometry and the rate gyro and eliminates the need for a robot equipped with wheel encoders.

\begin{figure}
    \centering
    \includegraphics[width=\columnwidth, trim={10cm 2cm 5cm 0cm}, clip]{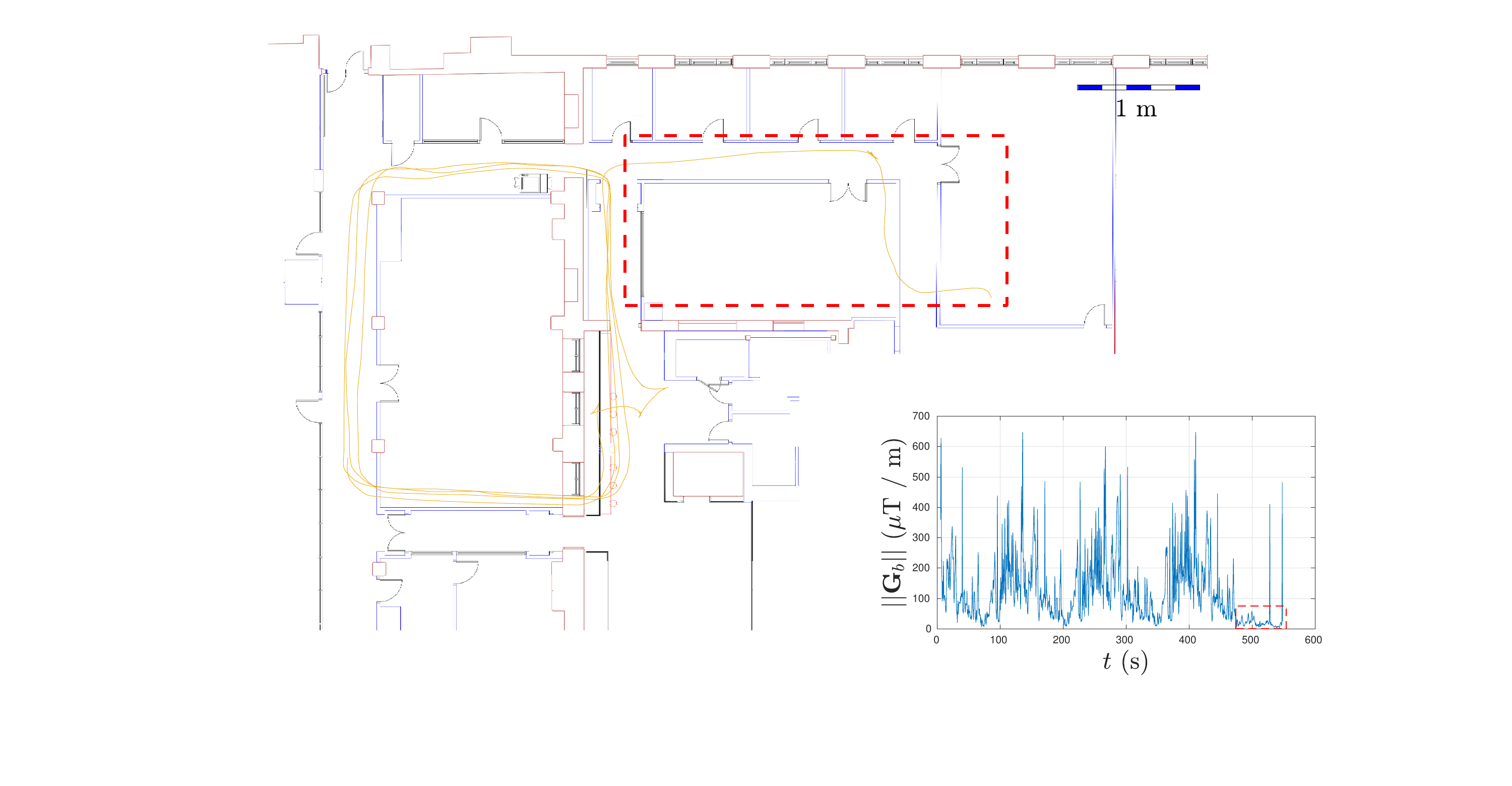} \vspace{-30pt}
    \caption{Top view of trajectory for Experiment 1 overlaid on map. The trajectory is estimated using the proposed solution with loop closures being enforced. The norm of the magnetic field gradient over time is shown in the bottom right corner.}
    \label{fig:birds_eye_map} \vspace{-0.7cm}
\end{figure}

The trajectory estimated using the proposed solution with loop closing is shown overlaid on a map of the space in which the experiment was performed in Figure~\ref{fig:birds_eye_map}. Because no ground truth was available in this experiment, the map is used to validate the algorithm. The alignment is performed visually, with no scaling applied. The map indicates that the error was approximately $1.5\, \SI{}{\meter}$. The norm of the magnetic field gradient for this trajectory is plotted in the bottom right of Figure~\ref{fig:birds_eye_map}. The largest error occurs in the portion of the trajectory encompassed by the red rectangle. The gradient measured during this portion of the trajectory is indicated by the red box in Figure~\ref{fig:birds_eye_map}. Here, it can be seen that, other than two large spikes in the gradient data, the norm of the magnetic field gradient has a smaller magnitude during this portion of the trajectory, varying from $0$ to $60 \, \frac{\mu\SI{}{\tesla}}{\SI{}{\meter}}$, rather than from $0$ to $700 \, \frac{\mu\SI{}{\tesla}}{\SI{}{\meter}}$ as it did during the rest of the experiment. This suggests that the algorithm performs best when the magnetic field gradient contains large variations.

It can be seen in Figure~\ref{fig:birds_eye_map} that the robot is estimated to have passed through a wall. While this indicates error in the estimates, the error is between 1 and $2\, \SI{}{\meter}$ over a trajectory in an area measuring $16 \times 27 \, \SI{}{\meter}^2$. Furthermore, the robot possessed no knowledge of the map, nor were any sensors used that were capable of sensing walls, such as LiDAR. Incorporating a prior map of the indoor space will be considered in the future, but is omitted here to simplify the presentation of the proposed navigation approach.

Note that the solution represented by the orange line in Figure~\ref{fig:birds_eye} illustrates that small errors can be achieved without enforcing loop closures. Although the robot trajectory involves laps of four different hallway sections, these laps are not required for adequate results to be obtained. In fact, in the solution represented by the orange line in Figure~\ref{fig:birds_eye}, the algorithm has no knowledge of the fact that it is visiting a previous location. Information pertaining to previously-visited locations is only used when loop closures are enforced.

\subsection{Experiment 2}

A second experiment is performed in a laboratory environment, in an area measuring $2.5 \times 4.5 \, \SI{}{\meter}^2$. The covariance of the pseudomeasurements, wheel slip and loop closures are varied slightly from those used in Experiment 1, to account for the difference in the calibration of the magnetometers. {These covariances are hand-tuned using the collected data.} A motion capture system is used to collect ground truth data.
The robot trajectory follows arbitrary motion about the space, making it unique from the trajectory of Experiment 1. During this experiment, the magnetic field measurements are collected with a frequency of $25\, \SI{}{\hertz}$. Loop closures are enforced in the reported results.

\begin{figure}
    \centering
    \includegraphics[width=0.9\columnwidth]{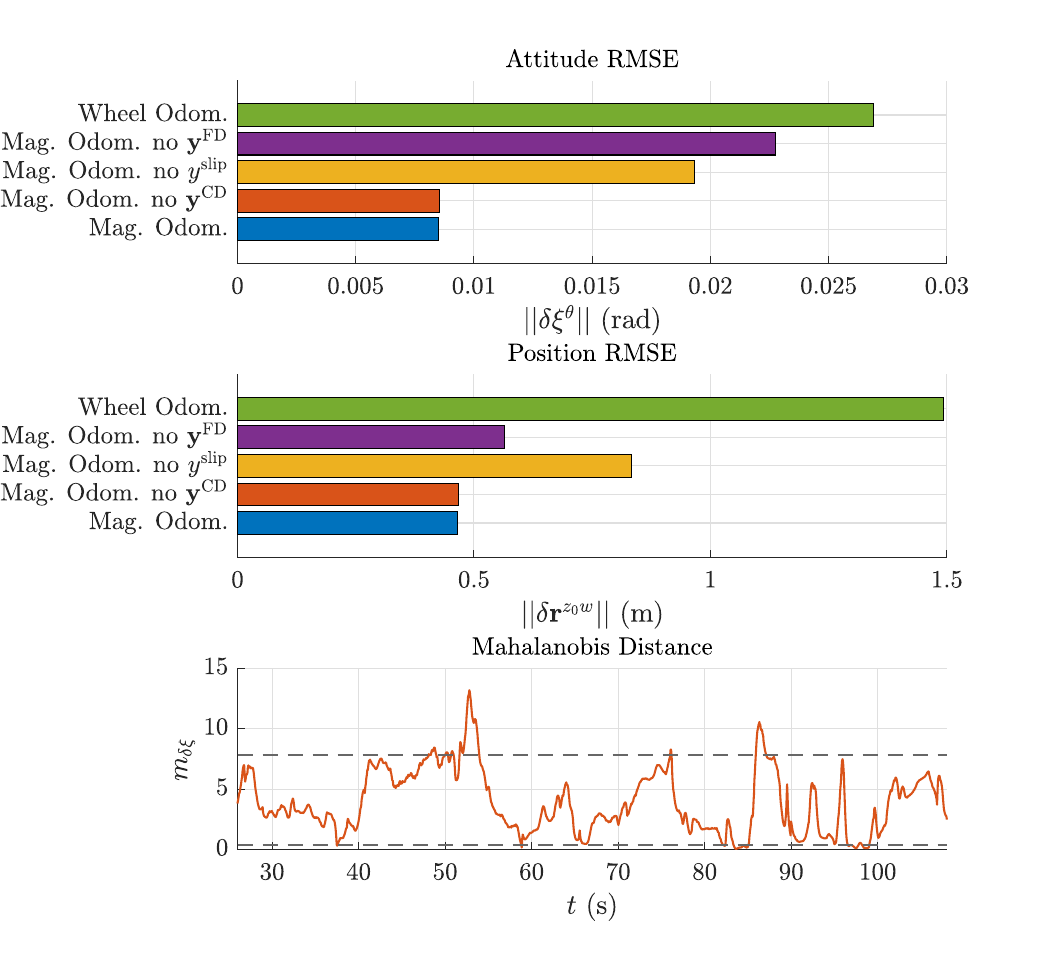}\vspace{-20pt}
    \caption{Top: Root mean squared error (RMSE) for Experiment 2. The RMSE of the trajectories estimated using magnetic odometry and wheel odometry are compared and results of an ablation study are presented. Bottom: Mahalanobis distance for Experiment 2. The dotted line represents the cumulative probability threshold corresponding to a significance level of 0.05.}
    \label{fig:lab_rmse} \vspace{-0.5cm}
\end{figure}

The performance of the algorithm is analysed through the root mean squared error (RMSE) of the attitude and position estimates, presented for magnetic odometry and wheel odometry in Figure~\ref{fig:lab_rmse}. The RMSE is computed as the mean of the error at each timestep, where the estimated trajectory is aligned to ground truth by fixing the first pose. While both magnetic odometry and wheel odometry result in small attitude estimation error, it is evident that magnetic odometry results in significant improvement in position estimation over the wheel odometry solution. The attitude RMSE for magnetic odometry is $0.0085\, \SI{}{\radian}$, which is an improvement upon the attitude RMSE of $0.0269\, \SI{}{\radian}$ achieved using wheel odometry. The position RMSE using wheel odometry is $1.492\, \SI{}{\meter}$, and using magnetic odometry is $0.566\, \SI{}{\meter}$. This corresponds to a 62\% improvement over the wheel odometry solution.

Mahalanobis distance is used to evaluate the consistency of the state estimation errors for the states estimated using magnetic odometry \cite{BarShalom2001}. The Mahalanobis distance plot in Figure~\ref{fig:lab_rmse} shows the Mahalanobis distance with a threshold corresponding to a significance level of 0.05 shown as the dashed line. This plot shows that the estimation errors are roughly consistent since the Mahalanobis distance is between the upper and lower bounds for the majority of the experiment.

An albation study is conducted by running the algorithm without~\eqref{eq:pseudo2},~\eqref{eq:pseudo3} and~\eqref{eq:slip}. RMSE for position and attitude are computed and shown in Figure~\ref{fig:lab_rmse}. Eliminating the finite difference measurement,~\eqref{eq:pseudo2}, leads to a 168\% increase in attitude RMSE and 21\% increase in position RMSE. The signifgicantly degraded attitude estimation performance shows that the finite difference measurement contributes a significant amount of information about the robot's attitude. Eliminating the use of the wheel slip pseudomeasurement,~\eqref{eq:slip}, results in a 128\% increase in attitude RMSE and a 79\% increase in position RMSE, which suggests that the wheel slip constraint significantly improves the estimates. 
Finally, eliminating the use of the central-difference-like pseudomeasurement,~\eqref{eq:pseudo3}, has little effect on the performance of the algorithm, increasing attitude RMSE by 0.7\% and position RMSE by 0.2\%. This indicates that the finite-difference-like pseudomeasurement given by~\eqref{eq:pseudo2} is sufficiently accurate to perform pose estimation. 
{ Note that while the central-difference-like pseudomeasurement does not have a significant effect on the errors in this experiment, it is expected to have a greater impact in cases in which the gradients have the biggest spatial change.}

\section{Conclusion and future work} \label{sec:conclusion}
This paper presents a novel method of performing navigation in indoor environments using magnetic field and magnetic field gradient measurements in conjunction with rate-gyro measurements. The magnetic field and its gradient are used to extract relative positions between poses. This admits the estimation of robot poses using only a rate gyro and an array of four magnetometers. Moreover, the magnetic field and magnetic field gradient are used to generate attitude-invariant components that can be used for loop closure detection. The proposed approach can be used alongside additional sensors within a navigation algorithm.
While the proposed navigation method was applied to magnetic fields, the derivations presented in Section~\ref{sec:MO} are applicable to any vector fields. This method could, for example, be applied to navigation using gravity fields \cite{Jircitano1991}.

\section*{Acknowledgment}
The authors thank Manon Kok and Arno Solin for sharing magnetic field data for testing in simulation.

{\AtNextBibliography{\small}
\balance
\printbibliography}

@phdthesis{Dorveaux2011,
  author  = {Dorveaux, Eric},
  title   = {Magneto-inertial navigation: principles and application to an indoor pedometer},
  school  = {École Nat. Supérieure des Mines de Paris},
  address = {Paris},
  year    = {2011},
  month   = {03}
}

@article{Brossard2020,  
author={Brossard, Martin and Barrau, Axel and Bonnabel, Silvère},  
journal={IEEE Trans. Intell. Veh.},   
title={{AI-IMU} Dead-Reckoning},   
year={2020},  
volume={5},  
number={4},  
pages={585-595}
}

@software{Ceres,
  author = {Agarwal, Sameer and Mierle, Keir and The Ceres Solver Team},
  title = {{Ceres Solver}},
  license = {Apache-2.0},
  url = {https://github.com/ceres-solver/ceres-solver},
  version = {2.1},
  year = {2022},
  month = {3}
}

@article{preint2017,
  author={Forster, Christian and Carlone, Luca and Dellaert, Frank and Scaramuzza, Davide},
  journal={IEEE Trans. Robot}, 
  title={On-Manifold Preintegration for Real-Time Visual--Inertial Odometry}, 
  year={2017},
  volume={33},
  number={1},
  pages={1-21}
}

@ARTICLE{paper_Sola_eta_al_micro,
  AUTHOR =       "Joan Sol\`{a} and Jeremie Deray and Dinesh Atchuthan",
  TITLE =        "{A micro Lie theory for state estimation in robotics}",
  JOURNAL =      "arXiv",
  year = {2020},
  volume =       {},
  number =       {},
  pages =        {},
  month =        "",
  note =         {arXiv:1812.01537},
  source =       {},
}

@inproceedings{Kok2018,
  author={Kok, Manon and Solin, Arno},
  booktitle={FUSION}, 
  title={Scalable Magnetic Field {SLAM} in {3D} Using {G}aussian Process Maps}, 
  year={2018},
  pages={1353-1360}
}

@phdthesis{Vallivaara2018,
  author       = {Vallivaara, Ilari}, 
  title        = {Simultaneous localization and mapping using the indoor magnetic field},
  school       = {University of Oulu},
  year         = 2018
}

@phdthesis{Mount2018,
  author       = {Mount, Lauren A.}, 
  title        = {Navigation using vector and tensor measurements of the {E}arth's magnetic anomaly field},
  school       = {Air Force Institute of Technology},
  year         = 2018
}

@inbook{Jung2019,
author = {Jung, Jongdae and Choi, Jinwoo and Oh, Taekjun and Myung, Hyun},
year = {2019},
month = {01},
booktitle = {RiTA},
pages = {153-163},
title = {Indoor Magnetic Pose Graph {SLAM} with Robust Back-End}
}

@inproceedings{Zmitri2019,
  author={Zmitri, Makia and Fourati, Hassen and Prieur, Christophe},
  booktitle={ Int. Conf. Indoor Position. Indoor Navig.}, 
  title={Improving Inertial Velocity Estimation Through Magnetic Field Gradient-based Extended {K}alman Filter}, 
  year={2019},
  pages={1-7}
}

@inproceedings{Vissiere2007,
  author={Vissiere, David and Martin, Alain and Petit, Nicolas},
  booktitle={Proc.~IEEE Conf. Decis. Control}, 
  title={Using distributed magnetometers to increase {IMU}-based velocity estimation into perturbed area}, 
  year={2007},
  pages={4924-4931},
}

@phdthesis{Canciani2016,
  author  = "Canciani, Aaron J.",
  title   = "Absolute positioning using the {E}arth's magnetic anomaly field",
  school  = "Air Force Institute for Technology",
  address = "Ohio, USA",
  year    = 2016,
  month   = 09
}

@Article{Viset2022,
AUTHOR = {Viset, Frida and Helmons, Rudy and Kok, Manon},
TITLE = {An Extended {K}alman Filter for Magnetic Field {SLAM} Using {G}aussian Process Regression},
JOURNAL = {Sensors},
VOLUME = {22},
YEAR = {2022},
NUMBER = {8},
ARTICLE-NUMBER = {2833},
URL = {https://www.mdpi.com/1424-8220/22/8/2833}
}

@inproceedings{Vallivaara2010,
  author={Vallivaara, Ilari and Haverinen, Janne and Kemppainen, Anssi and Röning, Juha},
  booktitle={IEEE Int. Conf. Multisens. Fusion Integr. Intell. Syst.}, 
  title={Simultaneous localization and mapping using ambient magnetic field}, 
  year={2010},
  pages={14-19}
}

@book{Purcell2012,
  author    = {Purcell, Edward M. and Morin, David J.}, 
  title     = {Electricity and Magnetism},
  publisher = {Cambridge University Press},
  year      = 2012,
  address   = {Cambridge, UK},
  edition   = 3
}

@article{Heath2003,
author = {Heath, Philip and Heinson, Graham and Greenhalgh, Stewart},
year = {2003},
month = {03},
title = {Some comments on potential field tensor data},
volume = {34},
journal = {Exploration Geophysics}
}

@book{Barfoot2017,
  author    = {Barfoot, Timothy D.}, 
  title     = {State estimation for robotics},
  publisher = {Cambridge University Press},
  year      = 2017
}

@article{Mangelson2020,
  author={Mangelson, Joshua G. and Ghaffari, Maani and Vasudevan, Ram and Eustice, Ryan M.},
  journal={IEEE Trans. Robot.}, 
  title={Characterizing the Uncertainty of Jointly Distributed Poses in the {L}ie Algebra}, 
  year={2020},
  volume={36},
  number={5},
  pages={1371-1388}
}

@article{Neira2001,
  author={Neira, J. and Tardos, J.D.},
  journal={IEEE Trans. Robot. Autom.}, 
  title={Data association in stochastic mapping using the joint compatibility test}, 
  year={2001},
  volume={17},
  number={6},
  pages={890-897}
}

@inproceedings{Robertson2013,
  author={Robertson, Patrick and Frassl, Martin and Angermann, Michael and Doniec, Marek and Julian, Brian J. and Garcia Puyol, Maria and Khider, Mohammed and Lichtenstern, Michael and Bruno, Luigi},
  booktitle={Int. Conf. Indoor Position. Indoor Navig.}, 
  title={Simultaneous Localization and Mapping for pedestrians using distortions of the local magnetic field intensity in large indoor environments}, 
  year={2013},
  pages={1-10}
}

@inproceedings{Frassl2013,
  author={Frassl, Martin and Angermann, Michael and Lichtenstern, Michael and Robertson, Patrick and Julian, Brian J. and Doniec, Marek},
  booktitle={IEEE Int. Conf. Intell. Robots Syst.}, 
  title={Magnetic maps of indoor environments for precise localization of legged and non-legged locomotion}, 
  year={2013},
  pages={913-920}
}

@book{BarShalom2001,
  author    = {Bar-Shalom, Yaakov and Li, X Rong and Thiagalingam, Kirubarajan}, 
  title     = {Estimation with Applications to Tracking and Navigation},
  publisher = {Wiley},
  year      = 2001
}

@misc{Twinleaf,
    author  = "Twinleaf",
    howpublished = "\url{https://twinleaf.com/vector/VMR/}",
    title         = "{VMR} Magnetoresistive Vector Magnetometer",
    note = "Accessed: September 14, 2021"
}

@INPROCEEDINGS{Jircitano1991,
  author = {{Jircitano}, Albert and {Dosch}, Daniel E.},
  title = "{Gravity aided inertial navigation system ({GAINS})}",
  booktitle = {Institute of Navigation, 47th Annual Meeting},
  year = 1991,
  volume = {47},
  month = {01},
  pages = {221-229}
}

@article{Lee2020,
author = {Lee, Taylor N. and Canciani, Aaron J.},
title = {{MagSLAM}: {A}erial simultaneous localization and mapping using {E}arth's magnetic anomaly field},
journal = {Navigation},
volume = {67},
number = {1},
pages = {95-107},
year = {2020}
}
\end{document}